\pdfoutput=1

\documentclass[11pt, anonymous]{article}
\usepackage[final]{acl}

\usepackage{times}
\usepackage{latexsym}
\usepackage{float}
\usepackage{graphicx}
\usepackage{float}
\usepackage{tabularx}
\usepackage{multirow} 
\usepackage{multicol} 
\usepackage{subfigure}
\usepackage{enumitem}

\usepackage[T1]{fontenc}

\usepackage[utf8]{inputenc}

\usepackage{microtype}

\usepackage{inconsolata}

\usepackage{enumitem}

\usepackage{graphicx}
\usepackage{listings}
\usepackage{amsmath}
\usepackage{amsthm}
\usepackage{amssymb}
\usepackage{adjustbox}
\usepackage{booktabs}

\usepackage{xcolor}

\definecolor{customgreen}{RGB}{0,168,62}  

\title{
 An empirical study of validating synthetic data for formula generation
}

\author{Usneek Singh \\ 
  Microsoft \\
  Bangalore, India \\
  \And
  Jos{\'e} Cambronero\thanks{Work done at Microsoft} \\ 
  Google \\ 
  Atlanta, USA \\
  \And
  Sumit Gulwani \\ 
  Microsoft \\ 
  Redmond, USA \\
  \AND
  Aditya Kanade \\ 
  Microsoft \\ 
  Bangalore, India \\
  \And
  Anirudh Khatry\footnotemark[1]\\ 
  University of Texas at Austin \\ 
  Austin, USA \\
  \And
  Vu Le \\ 
  Microsoft \\ 
  Redmond, USA \\
  \AND
  Mukul Singh \\ 
  Microsoft \\ 
  Redmond, USA \\
  \And
  Gust Verbruggen \\
  Microsoft \\ 
  Keerbergen, Belgium
  }

\begin{document}
\maketitle
\begin{abstract}
Large language models (LLMs) can be leveraged to help write formulas in spreadsheets, but formula data resources are scarce, impacting both the base performance of pre-trained models and limiting the ability to fine-tune them.
Given a corpus of formulas, we can use another model to generate synthetic natural language utterances for fine-tuning.
However, it is important to validate whether the natural language (NL) generated by the LLM is accurate for it to be beneficial for fine-tuning.
In this paper, we provide empirical results on the impact of validating these synthetic training examples with surrogate objectives that evaluate the accuracy of the synthetic annotations.
We demonstrate that validation improves performance over raw data across four models (2 open and 2 closed weight).
Interestingly, we show that although validation tends to prune more challenging examples, it increases the complexity of problems that models can solve after being fine-tuned on validated data.
\end{abstract}

\section{Introduction}
Derived-column formulas in spreadsheets generate a new column by transforming existing columns in a table, and they have been shown to be challenging to write \cite{gulwani2012spreadsheet}.
To aid users in writing such formulas, we can ask for a description in natural language \cite{zhao2024nl2formula}.
Unfortunately, since such formulas are sparse,  pre-trained language models (especially smaller) struggle in generating them without fine-tuning (for example, one of our models, Phi-2, achieved a pass@10 score of only 0.03, indicating a very low success rate in generating the correct formulas within 10 attempts.).

To construct a dataset for fine-tuning, public spreadsheet workbooks can be used but they contain only tables and formulas, whereas a fine-tuning dataset also requires paired natural language (NL) descriptions corresponding to each (Table, Formula). Traditionally datasets for NL-to-code tasks have been manually annotated \cite{zhou2024lima, austin2021program}. This is a time-consuming and expensive process. Leveraging LLMs, known for their text generation capabilities, is a viable alternative \cite{tan2024large} assuming that the \textbf{synthetic NL generated by LLMs is accurate}, as recent studies have shown that quality is more important than quantity \cite{zhou2024lima, li2023quantity, lozhkov2024fineweb-edu}.

In this paper, we leverage LLMs to predict the accuracy of synthetic NL using 3 surrogate objectives, and show empirical results of fine-tuning models on subsets of synthetic data that are accepted by these objectives.
Fine-tuning models on validated subsets shows better performance in predicting formulas compared to using raw data. For example, GPT-4 fine-tuned on data validated by generating code in an alternate common programming language saw up to a 28\% improvement in evaluation scores along with a 23\% reduction in training time. Additionally, we observe that the models fine-tuned on validated data perform better on more complex problems. We also find that models fine-tuned on validated data still manage to learn to use functions removed during validation. 

Our key contributions are as follows.
\begin{itemize}[itemsep=0pt, parsep=0pt,topsep=0pt]
    \item We define three surrogate objectives (output prediction, alternative code generation, and classification) to predict accuracy of synthetic natural language in the NL-to-Formula task.
    \item We empirically analyze the effect of validating synthetic data using these objectives on fine-tuning performance of different models.
\end{itemize}

\section{Related work}

\paragraph{Formula generation} 
FlashFill \cite{gulwani2011automating, gulwani2012spreadsheet} generates derived-column formulas by example, as users struggle with this task.
SpreadsheetCoder \cite{chen2021spreadsheetcoder} suggests formulas from surrounding context in spreadsheets. 
\textsc{flame} \cite{joshi2024flame} is a small language model that understands formulas for tasks like repair and retrieval, but does not handle natural language.
The NL-to-Formula (NL2F) task is introduced with a dataset obtained by converting the \textsc{Text2SQL} dataset to spreadsheet formulas \cite{zhao2024nl2formula}.
Unlike \cite{zhao2024nl2formula}, our work centers on empirically evaluating different NL validation strategies.

\paragraph{LLMs for synthetic data} 
\citet{tan2024large} discusses the applications of LLMs in data annotation for classification tasks. \citet{goel2023llms} demonstrates the use of LLMs in the medical domain, where they assist in labeling data with expert verification.
\citet{wang2024human}, \citet{kim2024meganno+}, and \citet{tang2024pdfchatannotator} explore human-LLM collaborative approaches for annotation and verification.
There has been no comparison of NL validation techniques on synthetic NL for NL2F.

\paragraph{Data quality for LLM fine-tuning}
\citet{chen2024automated} proposed an approach for automated filtering and verification of datasets to ensure high quality for LLM fine-tuning, leveraging the BSDetector \cite{chen2023quantifying} to obtain confidence scores from LLM outputs.
These techniques require existing ground truth labels (utterances) which are not available in our case. 
\citet{zhou2024lima} and \citet{li2023quantity} manually curate data to demonstrate that instruction tuning with a small (< 1000) set of high-quality examples yields competitive results. While their work focuses on selecting examples based on alignment (already assuming correctness), our work evaluates technique-based selection on accuracy of NL instructions.
\section{Validating synthetic data}

Let $T = [C_i]_1^n$ be a table with $n$ columns uniquely identified by a corresponding $h_i$ label.
A derived-column formula $F$ is a formula where each leaf node in the AST (Abstract Syntax Tree) of $F$ is either a constant value or a column identifier $h_i$.
Let $U$ be an utterance in natural language that describes how to derive a column from $T$.
A derived-column task is specified by $(U, T, F)$.
Given $U$ and $T$ the goal is to find a formula $F^\prime$ such that $F^\prime(T) \equiv F(T)$, where equivalence indicates both formulas produce the same outputs given the same inputs.

To fine-tune a model, we therefore need examples of the form $(U, T, F)$.
$T$ and $F$ can be mined from large spreadsheet corpora \cite{cornet,joshi2024flame} and we can use an LLM to generate an utterance $\hat{U} = LLM(T, F)$.

\begin{figure}[t] 
    \centering
    \includegraphics[width=1.05\columnwidth]{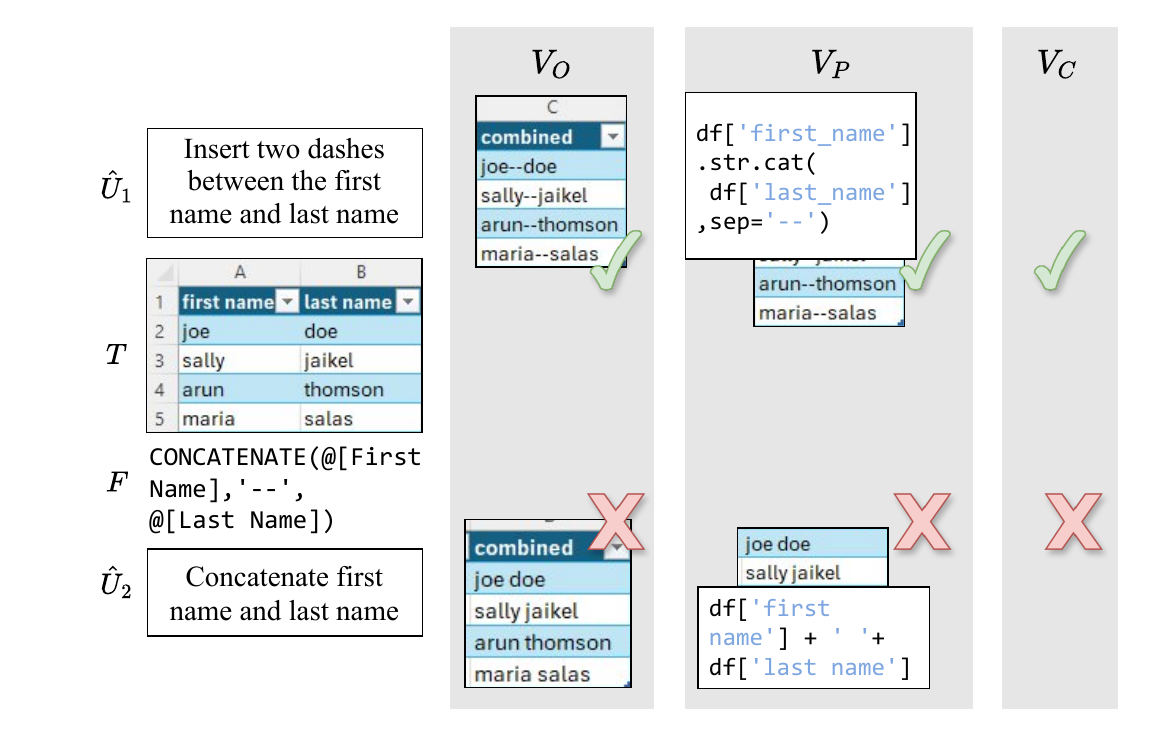}
    \caption{Overview of different validators implemented on top of GPT-4 represented by (a) \textbf{\boldmath$V_O$}: This validator directly computes $F(T)$ from $(\hat{U},T)$; (b) \textbf{\boldmath$V_P$}: Validator predicts python program $P$ from $(\hat{U},T)$ to compare $P(T)$ with $F(T)$; (c) \textbf{\boldmath$V_C$}: Validator directly classifies $\hat{U}$ based on input $(\hat{U}, T, F)$.}

   
    \label{fig:validation}
    \vspace{-0.2cm}
\end{figure}

A \emph{validator} $V(\hat{U}, T, F) \rightarrow \mathbb{B}$ is a function that predicts whether $\hat{U}$ accurately describes the formula $F$ operating on table $T$.
These validators can be defined in any way---even using human annotators.
To reduce manual effort, we define three validators using an LLM.
An overview of these three validators is shown in Figure~\ref{fig:validation}.

\paragraph{Output prediction ($V_O$)}

This validator asks the LLM to directly predict the output values $F(T)$ from $(\hat{U},T)$ and uses an element-wise row comparison to evaluate correctness. 
For numbers, we allow an absolute difference of 0.05.
For strings, we use a longest common sub-sequence ratio of 0.8 as passing criterion. 
This approach leverages natural language to emulate the computation directly. It is inspired from the alternate task of output prediction discussed in \citet{khatry2023words}
\begin{figure*}[t] 
    \centering
    \includegraphics[width=\textwidth]{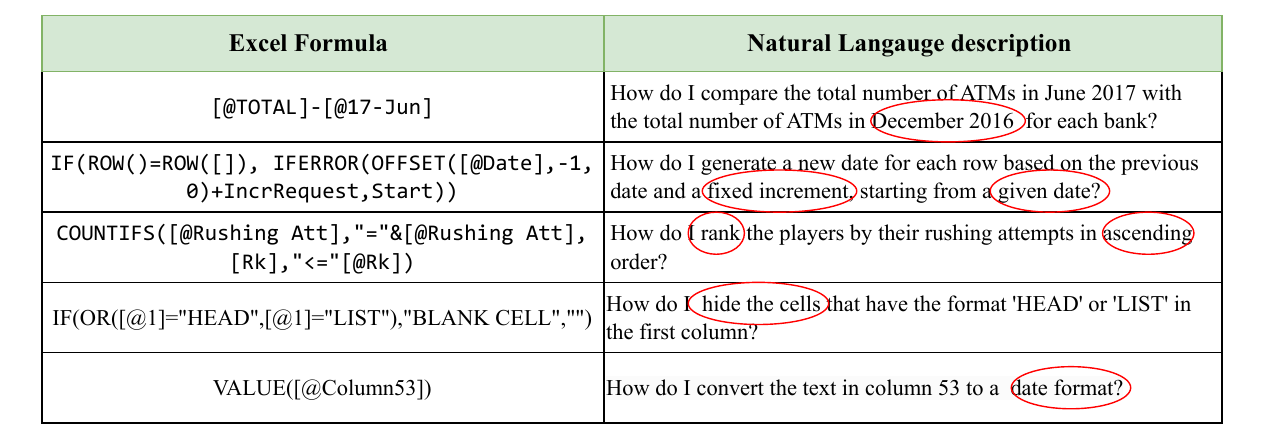}
    \caption{Examples of cases filtered by validators implemented on top of GPT-4. The synthetic natural language descriptions in these examples are under-specified, contain incorrect intent, or convey an unclear idea.}

   
    \label{fig:examples}
    \vspace{-0.2cm}
\end{figure*}

\paragraph{Alternate code generation ($V_P$)}

This validator asks the LLM to predict a program $P$ in another language (we use Python) from $(\hat{U},T)$ and compares $P(T)$ (execution of P on $T$) with $F(T)$ using element-wise comparison with the same relaxations for strings and numbers previously described. This leverages the abilities of LLMs to generate popular programming languages  \cite{10.5555/3618408.3619494}.

\paragraph{Classification ($V_C$)}
This validator directly asks the model to classify whether $\hat{U}$ accurately describes $F$ over $T$.
It is based on the self-reflection certainty objective from BSDetector \cite{chen2023quantifying}.

More details about the validators are provided in Appendix \ref{subsec:details}. 
We also provide a few examples in Figure \ref{fig:examples} to illustrate cases filtered by the validators from the raw dataset.

\section{Experimental setup}
We describe training data and models, and the testing benchmark.

\paragraph{Training data}
We mine $(T, F)$ pairs that satisfy our derived-column definition from publicly available Excel workbooks \cite{cornet}. 
We create a training set and validation set of size 7833 and 422 respectively. Each $(T, F)$ pair is annotated with an utterance $\hat{U}$ using GPT-4 at a low temperature.

\paragraph{Models}

\newcommand{\phitwo}[0]{\texttt{\small phi-2}}
\newcommand{\mistral}[0]{\texttt{\small mistral}}
\newcommand{\gptt}[0]{\texttt{\small gpt-35}}
\newcommand{\gptf}[0]{\texttt{\small gpt-4}}

We use two open (\phitwo{} (2B) and \mistral\texttt{\small -7b-instruct} (7B)) and two closed-weight (\gptt\texttt{\small-turbo} and \gptf{}) models.
\phitwo{} (8 $\times$ V100) and \mistral{} (1 $\times$ A100) were fine-tuned for 10 and 15 epochs respectively. We selected the best checkpoint using validation loss.
\gptt{} ($16$ $\times$ A100) and \gptf{} ($24$ $\times$ A100) were fine-tuned using the Azure API.
\mistral{}, \gptt{}, \gptf{} were fine-tuned using LoRA \cite{hu2021lora}.


\paragraph{Testing data}



The \textsc{SofSet} dataset \cite{barke2024solving} consists of 201 spreadsheet formula tasks from StackOverflow.
Of these, we filter the 139 tasks that satisfy our derived-column definition.

\paragraph{Metric}

We use the pass@$k$ metric \cite{chen2021evaluating} based on execution match of formula, were $k$ represents the number of predictions considered out of the total number of predictions provided.
In our evaluation system, we generate $n=10$ predictions at temperature $0.6$ and compute pass@5 metric.
\section{Results and Discussion}

We perform experiments to empirically explore the following research questions.

\begin{itemize}[itemsep=0pt, parsep=0pt,topsep=0pt,align=left]
    \item[\bfseries RQ1] How do different validators compare?
    \item[\bfseries RQ2] What is the impact of validating data on fine-tuning performance?
    \item[\bfseries RQ3] What are the differences in cases solved by models trained on validated NL and raw dataset? 
    \item[\bfseries RQ4] Can models finetuned on validated data learn the functions removed during validation? 
\end{itemize}

\subsection{RQ1: Comparing validators} \label{subsec:NL-validation}

We apply our three validation approaches to our initial set of 7833 points. This produces the data subsets described in Table~\ref{tab:dataset_characteristics}.
We shows properties of the formulas accepted by each validator.
Since $V_O$ is bottle-necked on numerical operations, it succeeds for fewer unique functions and operators.
Similarly, $V_P$ struggles with more functions than $V_C$ as there might not be an easy Python equivalent.
\begin{table}[h]
    \centering
    \caption{Summary of training data subsets with different validation approaches. "\# functions" refers to unique functions, "\# calls" to average function calls, "depth" to function nesting level, and "\# ops" to average arithmetic operator count in formulas.} 
    \small
    \label{tab:dataset_characteristics}
    \begin{tabular}{lrrrrr}
        \toprule
        $V$ & \textbf{Size} & \textbf{\# functions} & \textbf{\# calls} & \textbf{depth} & \textbf{\# ops} \\ \midrule
        $\varnothing$ & 7833 & 122 & 1.03 & 0.87 & 1.28 \\ 
        $V_O$ & 2266 & 71 & 0.71 & 0.65 & 1.01 \\ 
        $V_P$ & 4095 & 95 & 0.86 & 0.77 & 1.22 \\ 
        $V_C$ & 5246 & 109 & 0.87 & 0.79 & 1.24 \\
        \bottomrule
    \end{tabular}
\end{table}

\begin{figure}[h] 
    \centering
    \includegraphics[width=0.5\columnwidth]{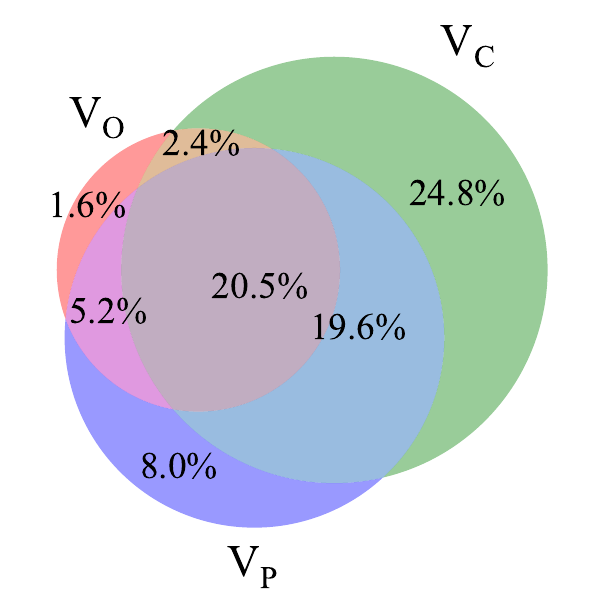} 
    \caption{Summary of overlap of different data subsets produced by different validation strategies.}
    \label{fig:venn}
    \vspace{-0.5cm}
\end{figure}

Figure~\ref{fig:venn} shows overlap in examples accepted by different validators.
Each validator uniquely accepts at least some examples.
1403 (18\%) examples does not pass any validator.

\subsection{RQ2: Effect on fine-tuning performance}


We compare the impact of validated data versus raw (unvalidated) data, as well as the impact of validated data versus rejected cases by each validator, on the downstream performance of the NL2F task.

\paragraph{Versus raw}

Table \ref{tab:benchmark2} shows base model (few-shot) and fine-tuning performance on different subsets of data.
For the smaller models, \phitwo{} and \mistral{}, the performance increase with fine-tuning is more significant.
With all models, a smaller, validated dataset yields better performance than raw data.
\textbf{\boldmath$V_P$ yields the best performance on average with nearly half the size of raw data.} \gptf{} improves only when fine-tuned on validated data. Surprisingly, \gptt{} without fine-tuning outperforms the fine-tuned version, likely due to differences in data distribution between training and testing benchmarks. 
Besides performance, fine-tuning with validated data also reduces training time significantly, as shown in Table~\ref{tab:training_time}. 
We see the performance on dataset created by the intersection of all validators (marked by $\cap$) is limited by the worst performing validator in each case.
\begin{table}[h]
\centering
\caption{Performance comparison of the different models on \textsc{SofSet} Benchmark using pass@5 metric. Three out of the four models give best performance when fine-tuned on data validated by $V_P$.}
\small
\label{tab:benchmark2}
\begin{tabular}[width=\textwidth]{lrrrrr}
\toprule
FT on & \# ex & \phitwo{} & \mistral{} & \gptt{} & \gptf{} \\
\midrule
Base & 0 & 0.02 & 0.04 & \textbf{0.30} & 0.36 \\ 
Raw & 7833 & 0.08 & 0.16 & 0.29 & 0.32 \\
$V_O$ & 2266 & 0.06 & \textbf{0.17} & 0.28 & 0.35 \\ 
$V_P$ & 4095 & \textbf{0.10} & 0.14 & \textbf{0.30} & \textbf{0.41} \\
$V_C$ & 5246 & 0.09 & 0.14 & \textbf{0.30} & 0.40 \\ 
$\cap$ & 1607 & 0.06 & 0.13 & 0.27 & 0.34 \\ 
\bottomrule
\end{tabular}
\end{table}

\begin{table}[h]
    \centering
    \caption{Training time and relative improvement for different models on data subsets. Models fine-tuned on $V_P$ and $V_C$ subsets require less time than on raw data while delivering better downstream performance. }
    \label{tab:training_time}
    \small
    \begin{tabular}[width=\textwidth]{lrrrr}
        \toprule
        \textbf{Data} & \phitwo{} &  \mistral{} & \gptt{} & \gptf{} \\
\midrule
Raw  & 15h44m & 8h51m & 4h45m & 14h00m \\
$V_O$ & -73\% & -71\% & -60\% & -47\% \\
$V_P$ & -48\% & -45\% & -37\% & -23\% \\
$V_C$ & -36\% & -32\% & -19\% & -19\% \\
\bottomrule
    \end{tabular}
    \vspace{-0.2cm}
\end{table}

\paragraph{Versus invalidated}

Table~\ref{tab:accept/reject} compares the performance of fine-tuning on the accepted (\& subsampled) and rejected ($\neg$) examples for each validator.
We sub-sample the accepted sets to 2266---the number of examples in the smallest set ($V_O$).
We observe that, despite the smaller size of the validated data subset (subsampled), it outperforms its larger invalidated (rejected) counterpart in most (11/12) comparisons.
The only case where this not happens is for $V_O$ on \gptf{}, likely due to the many functions (51) that were eliminated from the training data.
\begin{table}[h]
\centering
\caption{Pairwise comparison of performance of sub-sampled ($\subset$) data from validated ($V$) against rejected ($\neg V$) examples.
Results of pairs $(\subset V, \neg V)$ are marked in \textcolor{customgreen}{green} if $(\subset V > \neg V)$, \textcolor{blue}{blue} if $(\subset V = \neg V)$.
}
\small
\label{tab:accept/reject}
\begin{tabular}[width=\textwidth]{rrrrrr}
\toprule
FT on & \# ex & \phitwo{} & \mistral{} & \gptt{} & \gptf{} \\
\midrule
$V_O$ & 2266 & \textcolor{customgreen}{0.06}& \textcolor{customgreen}{0.17} & \textcolor{customgreen}{0.28} & \textcolor{blue}{0.35} \\ 
$\neg V_O$ & 5567 & \textcolor{customgreen}{0.05} & \textcolor{customgreen}{0.16} & \textcolor{customgreen}{0.27} &	\textcolor{blue}{0.35} \\ \midrule

$\subset{V_C}$  & 2266 & \textcolor{customgreen}{0.07} & \textcolor{customgreen}{0.14} & \textcolor{customgreen}{0.27} & \textcolor{customgreen}{0.36} \\ 
$\neg V_C$ & 2587 & \textcolor{customgreen}{0.04} &	\textcolor{customgreen}{0.12} & \textcolor{customgreen}{0.26} & \textcolor{customgreen}{0.34}\\ \midrule

$\subset V_P$ & 2266 & \textcolor{customgreen}{0.08}	& \textcolor{customgreen}{0.12} &	\textcolor{customgreen}{0.31} &	\textcolor{customgreen}{0.37} \\ 
$\neg V_P$ & 3738 & \textcolor{customgreen}{0.05} &	\textcolor{customgreen}{0.11} & \textcolor{customgreen}{0.24} & \textcolor{customgreen}{0.32} \\ \bottomrule
\end{tabular}
\end{table}




\subsection{RQ3: Analysing solved cases}
\begin{figure}[h]
    \centering
    \includegraphics[width=\columnwidth]{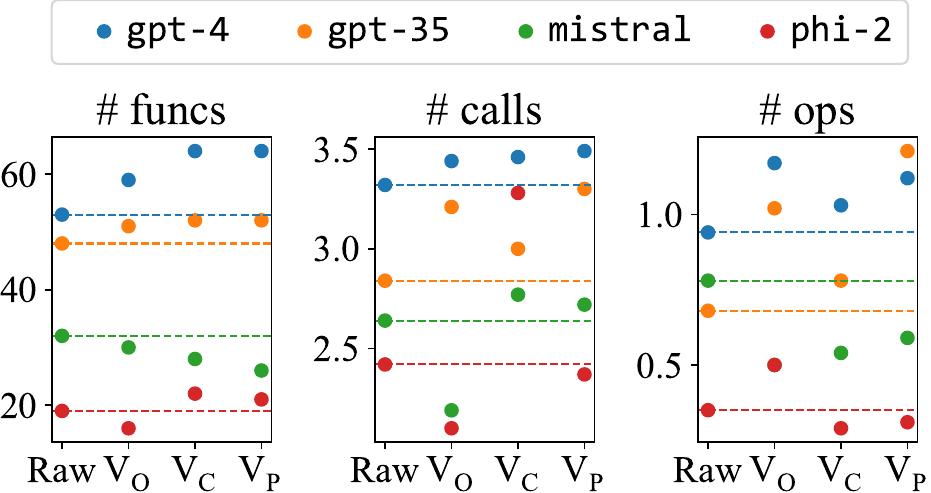}
    \caption{Comparison of correctly solved cases on models fine-tuned with different validation subsets based on (a) Number of unique functions (b) Average number of function calls (c) Average operator count of formulas}
    \label{fig:analysis}
    \vspace{-0.2cm}
\end{figure}

Figure~\ref{fig:analysis} shows properties of the solved cases (where at least one prediction was correct) after fine-tuning different models on raw data and validated subsets.
We see that fine-tuning on datasets with fewer unique functions still enables all models (except for \mistral{}) to solve cases with more unique functions.  
The average function call count increases for validated subsets compared to the raw data, indicating more complex formulas are solved by models fine-tuned on validated data. 
For \gptf{} and \gptt{}, average operator count also increases with fine-tuning on validated data.


\subsection{RQ4: Recovery of functions removed during validation}
By analyzing the output generations of the fine-tuned models, we identify new functions that were not present in their base model (without fine-tuning) predictions. Our results show that there are functions that were not in the fine-tuning dataset *and* were not in the base model's predictions, but after finetuning on validated datasets, we see these functions used in trained model predictions (see Table \ref{tab:new_functions}). 
This suggests that fine-tuning on a high-quality dataset allows the model to remember knowledge that it had learned during pre-training, without teaching it to hallucinate on potential mistakes in the synthetic data.

\begin{table}[h]
    \centering
    \caption{Number of functions learned by different fine-tuned models that were removed during validation. Some examples include {'SUMIF', 'TIME', 'QUOTIENT','ROWS','AGGREGATE'}}
    \label{tab:new_functions}
    \small
    \begin{tabular}{lrrrr}
        \toprule
        & \phitwo{} & \mistral{} & \gptt{} & \gptf{} \\
        \midrule
        \(V_0\) & 1 & 2 & 2 & 3 \\
        \(V_P\) & 3 & 2 & 3 & 3 \\
        \(V_C\) & 1 & 1 & 2 & 2 \\
        \bottomrule
    \end{tabular}
    \vspace{-0.2cm}
\end{table}

\subsection{Recommendations}
From our study, we see that a single validator (Alternate Code generation) works best on 3 out of 4 models. We also see that Output Prediction validator shows lower performance in general, likely because many functions (51) were removed from the training data during validation. However, practitioners should experiment with different validation methods, starting with the subset here, as validation in general improves performance.
\section{Conclusion}

We empirically evaluate the effect of automated validation of synthetic data using LLMs on the fine-tuning performance of derived-column NL-to-formula.
We validate synthetic NL annotations with three surrogate tasks (classification, code generation in Python, and output prediction) and fine-tune different models on the examples accepted by each of these methods.
In general, fine-tuning on smaller, validated datasets improves performance.
Despite validation resulting in datasets with simpler formulas, that does not cause the fine-tuned models to only solve simpler problems. 
Models fine-tuned on validated data are able to recover some functions that were removed during validation.

\section{Limitations}


Although we have focused on validating the correctness of natural language instructions, we have not addressed techniques for correcting them. Exploring methods for correcting instructions could be beneficial, as it would prevent the loss of data points. While having a smaller set of high-quality data can be advantageous for efficient training, achieving the best results may require maintaining a larger dataset by correcting invalid instructions.

In our study, the distribution of training data for fine-tuning is different than the testing data, which might not fully reflect the potential of fine-tuning. Additionally, our research has concentrated on formulas that expect a single, well-structured (formatted) input table. We aim to extend our work to include formulas that involve multiple tables and unstructured input. Furthermore, we have explored the potential of our technique in one language (English). We believe it will be valuable to investigate multilingual systems for validation setups.

\clearpage

\bibliography{custom}

\begin{thebibliography}{22}
\providecommand{\natexlab}[1]{#1}

\bibitem[{Austin et~al.(2021)Austin, Odena, Nye, Bosma, Michalewski, Dohan, Jiang, Cai, Terry, Le et~al.}]{austin2021program}
Jacob Austin, Augustus Odena, Maxwell Nye, Maarten Bosma, Henryk Michalewski, David Dohan, Ellen Jiang, Carrie Cai, Michael Terry, Quoc Le, et~al. 2021.
\newblock Program synthesis with large language models.
\newblock \emph{arXiv preprint arXiv:2108.07732}.

\bibitem[{Barke et~al.(2024)Barke, Poelitz, Negreanu, Zorn, Cambronero, Gordon, Le, Nouri, Polikarpova, Sarkar et~al.}]{barke2024solving}
Shraddha Barke, Christian Poelitz, Carina~Suzana Negreanu, Benjamin Zorn, Jos{\'e} Cambronero, Andrew~D Gordon, Vu~Le, Elnaz Nouri, Nadia Polikarpova, Advait Sarkar, et~al. 2024.
\newblock Solving data-centric tasks using large language models.
\newblock \emph{arXiv preprint arXiv:2402.11734}.

\bibitem[{Chen and Mueller(2023)}]{chen2023quantifying}
Jiuhai Chen and Jonas Mueller. 2023.
\newblock Quantifying uncertainty in answers from any language model and enhancing their trustworthiness.

\bibitem[{Chen and Mueller(2024)}]{chen2024automated}
Jiuhai Chen and Jonas Mueller. 2024.
\newblock Automated data curation for robust language model fine-tuning.
\newblock \emph{arXiv preprint arXiv:2403.12776}.

\bibitem[{Chen et~al.(2021{\natexlab{a}})Chen, Tworek, Jun, Yuan, Pinto, Kaplan, Edwards, Burda, Joseph, Brockman et~al.}]{chen2021evaluating}
Mark Chen, Jerry Tworek, Heewoo Jun, Qiming Yuan, Henrique Ponde de~Oliveira Pinto, Jared Kaplan, Harri Edwards, Yuri Burda, Nicholas Joseph, Greg Brockman, et~al. 2021{\natexlab{a}}.
\newblock Evaluating large language models trained on code.
\newblock \emph{arXiv preprint arXiv:2107.03374}.

\bibitem[{Chen et~al.(2021{\natexlab{b}})Chen, Maniatis, Singh, Sutton, Dai, Lin, and Zhou}]{chen2021spreadsheetcoder}
Xinyun Chen, Petros Maniatis, Rishabh Singh, Charles Sutton, Hanjun Dai, Max Lin, and Denny Zhou. 2021{\natexlab{b}}.
\newblock Spreadsheetcoder: Formula prediction from semi-structured context.
\newblock In \emph{International Conference on Machine Learning}, pages 1661--1672. PMLR.

\bibitem[{Goel et~al.(2023)Goel, Gueta, Gilon, Liu, Erell, Nguyen, Hao, Jaber, Reddy, Kartha et~al.}]{goel2023llms}
Akshay Goel, Almog Gueta, Omry Gilon, Chang Liu, Sofia Erell, Lan~Huong Nguyen, Xiaohong Hao, Bolous Jaber, Shashir Reddy, Rupesh Kartha, et~al. 2023.
\newblock Llms accelerate annotation for medical information extraction.
\newblock In \emph{Machine Learning for Health (ML4H)}, pages 82--100. PMLR.

\bibitem[{Gulwani(2011)}]{gulwani2011automating}
Sumit Gulwani. 2011.
\newblock Automating string processing in spreadsheets using input-output examples.
\newblock \emph{ACM Sigplan Notices}, 46(1):317--330.

\bibitem[{Gulwani et~al.(2012)Gulwani, Harris, and Singh}]{gulwani2012spreadsheet}
Sumit Gulwani, William~R Harris, and Rishabh Singh. 2012.
\newblock Spreadsheet data manipulation using examples.
\newblock \emph{Communications of the ACM}, 55(8):97--105.

\bibitem[{Hu et~al.(2021)Hu, Shen, Wallis, Allen-Zhu, Li, Wang, Wang, and Chen}]{hu2021lora}
Edward~J Hu, Yelong Shen, Phillip Wallis, Zeyuan Allen-Zhu, Yuanzhi Li, Shean Wang, Lu~Wang, and Weizhu Chen. 2021.
\newblock Lora: Low-rank adaptation of large language models.
\newblock \emph{arXiv preprint arXiv:2106.09685}.

\bibitem[{Joshi et~al.(2024)Joshi, Ebenezer, Sanchez, Gulwani, Kanade, Le, Radi{\v{c}}ek, and Verbruggen}]{joshi2024flame}
Harshit Joshi, Abishai Ebenezer, Jos{\'e}~Cambronero Sanchez, Sumit Gulwani, Aditya Kanade, Vu~Le, Ivan Radi{\v{c}}ek, and Gust Verbruggen. 2024.
\newblock Flame: A small language model for spreadsheet formulas.
\newblock In \emph{Proceedings of the AAAI Conference on Artificial Intelligence}, volume~38, pages 12995--13003.

\bibitem[{Khatry et~al.(2023)Khatry, Cahoon, Henkel, Deep, Emani, Floratou, Gulwani, Le, Raza, Shi, Singh, and Tiwari}]{khatry2023words}
Anirudh Khatry, Joyce Cahoon, Jordan Henkel, Shaleen Deep, Venkatesh Emani, Avrilia Floratou, Sumit Gulwani, Vu~Le, Mohammad Raza, Sherry Shi, Mukul Singh, and Ashish Tiwari. 2023.
\newblock \href {https://arxiv.org/abs/2305.01598} {From words to code: Harnessing data for program synthesis from natural language}.
\newblock \emph{Preprint}, arXiv:2305.01598.

\bibitem[{Kim et~al.(2024)Kim, Mitra, Chen, Rahman, and Zhang}]{kim2024meganno+}
Hannah Kim, Kushan Mitra, Rafael~Li Chen, Sajjadur Rahman, and Dan Zhang. 2024.
\newblock Meganno+: A human-llm collaborative annotation system.
\newblock \emph{arXiv preprint arXiv:2402.18050}.

\bibitem[{Li et~al.(2023)Li, Zhang, Li, Chen, Chen, Cheng, Wang, Zhou, and Xiao}]{li2023quantity}
Ming Li, Yong Zhang, Zhitao Li, Jiuhai Chen, Lichang Chen, Ning Cheng, Jianzong Wang, Tianyi Zhou, and Jing Xiao. 2023.
\newblock From quantity to quality: Boosting llm performance with self-guided data selection for instruction tuning.
\newblock \emph{arXiv preprint arXiv:2308.12032}.

\bibitem[{Lozhkov et~al.(2024)Lozhkov, Ben~Allal, von Werra, and Wolf}]{lozhkov2024fineweb-edu}
Anton Lozhkov, Loubna Ben~Allal, Leandro von Werra, and Thomas Wolf. 2024.
\newblock \href {https://huggingface.co/datasets/HuggingFaceFW/fineweb-edu} {Fineweb-edu}.

\bibitem[{Ni et~al.(2023)Ni, Iyer, Radev, Stoyanov, Yih, Wang, and Lin}]{10.5555/3618408.3619494}
Ansong Ni, Srini Iyer, Dragomir Radev, Ves Stoyanov, Wen-tau Yih, Sida~I. Wang, and Xi~Victoria Lin. 2023.
\newblock Lever: learning to verify language-to-code generation with execution.
\newblock In \emph{Proceedings of the 40th International Conference on Machine Learning}, ICML'23. JMLR.org.

\bibitem[{Singh et~al.(2023)Singh, S{\'a}nchez, Gulwani, Le, Negreanu, Raza, and Verbruggen}]{cornet}
Mukul Singh, Jos{\'e}~Cambronero S{\'a}nchez, Sumit Gulwani, Vu~Le, Carina Negreanu, Mohammad Raza, and Gust Verbruggen. 2023.
\newblock Cornet: Learning table formatting rules by example.
\newblock \emph{Proceedings of the VLDB Endowment}, 16(10):2632--2644.

\bibitem[{Tan et~al.(2024)Tan, Beigi, Wang, Guo, Bhattacharjee, Jiang, Karami, Li, Cheng, and Liu}]{tan2024large}
Zhen Tan, Alimohammad Beigi, Song Wang, Ruocheng Guo, Amrita Bhattacharjee, Bohan Jiang, Mansooreh Karami, Jundong Li, Lu~Cheng, and Huan Liu. 2024.
\newblock Large language models for data annotation: A survey.
\newblock \emph{arXiv preprint arXiv:2402.13446}.

\bibitem[{Tang et~al.(2024)Tang, Chang, and Yang}]{tang2024pdfchatannotator}
Yi~Tang, Chia-Ming Chang, and Xi~Yang. 2024.
\newblock Pdfchatannotator: A human-llm collaborative multi-modal data annotation tool for pdf-format catalogs.
\newblock In \emph{Proceedings of the 29th International Conference on Intelligent User Interfaces}, pages 419--430.

\bibitem[{Wang et~al.(2024)Wang, Kim, Rahman, Mitra, and Miao}]{wang2024human}
Xinru Wang, Hannah Kim, Sajjadur Rahman, Kushan Mitra, and Zhengjie Miao. 2024.
\newblock Human-llm collaborative annotation through effective verification of llm labels.
\newblock In \emph{Proceedings of the CHI Conference on Human Factors in Computing Systems}, pages 1--21.

\bibitem[{Zhao et~al.(2024)Zhao, Hou, Wu, Gao, Dong, Wan, Zhang, Sui, and Zhang}]{zhao2024nl2formula}
Wei Zhao, Zhitao Hou, Siyuan Wu, Yan Gao, Haoyu Dong, Yao Wan, Hongyu Zhang, Yulei Sui, and Haidong Zhang. 2024.
\newblock Nl2formula: Generating spreadsheet formulas from natural language queries.
\newblock \emph{arXiv preprint arXiv:2402.14853}.

\bibitem[{Zhou et~al.(2024)Zhou, Liu, Xu, Iyer, Sun, Mao, Ma, Efrat, Yu, Yu et~al.}]{zhou2024lima}
Chunting Zhou, Pengfei Liu, Puxin Xu, Srinivasan Iyer, Jiao Sun, Yuning Mao, Xuezhe Ma, Avia Efrat, Ping Yu, Lili Yu, et~al. 2024.
\newblock Lima: Less is more for alignment.
\newblock \emph{Advances in Neural Information Processing Systems}, 36.

\end{thebibliography}

\clearpage
\appendix
\label{sec:appendix}

\section{Training Data Characteristics} \label{subsec:training_data}
In this section, we summarise important formula properties for the training data extracted from excel workbooks (see Table \ref{tab:function_call_distribution}). From the original corpus, we remove any formulas that have deprecated functions to produce a set of 10,389 (table, formula) pairs. We then remove any pairs where the formula results in a missing/empty value for all output rows or uses multiple tables.
After the process of filtering, our final dataset consists of 7,833 (table, formula) pairs. This dataset has formulas which use 122 distinct
built-in functions. 
The most popular
functions match those typically
employed by Excel spreadsheet
users: \texttt{IF, SUM, IFERROR,
CONCATENATE, AND}. The other properties are summarised in Table \ref{tab:function_call_distribution}). The function call count refers to the frequency of Excel function calls within a formula. The depth of formulas denotes the extent of nested function calls within them. Operator count is the number of arithmetic operators (+, -, *, /) in a formula.
\begin{table}[h]
    \centering
    \caption{Characteristics of formulas used in Training Data obtained from Excel spreadsheets}
    \label{tab:function_call_distribution}
    \small
    \begin{tabular}{cccc}
        \toprule 
        & Fxn. call count & Formula depth & Op. count \\ \midrule
        0 & 3554 & 3554 & 2887 \\ 
        1 & 2625 & 2682 & 2811 \\ 
        2 & 965 & 1030 & 1169 \\ 
        3 & 285 & 325 & 435 \\ 
        4 & 115 & 125 & 187 \\ 
        $\geq5$ & 289 & 113 & 344 \\ \bottomrule
    \end{tabular}
\end{table}

\section{Model hyper-parameters used while Fine-tuning} \label{subsec:hyperparameters}
\paragraph{Phi-2}
For the Phi-2 model, fine-tuning was performed for 10 epochs with a batch size of 8. The learning rate was set to 1e-6, and the Adam optimizer was used along with a cross-entropy loss function.

\paragraph{Mistral}
The Mistral model was fine-tuned for 15 epochs using the LoRA technique \cite{hu2021lora}. The specific parameters for LoRA included a LoRA rank ($Lora\_r$) of 64, a LoRA alpha ($Lora\_alpha$) of 16, and a LoRA dropout ($Lora\_dropout$) of 0.1. The target modules for LoRA adaptation were "q\_proj", "k\_proj", "v\_proj", "o\_proj", "gate\_proj", "up\_proj", "down\_proj", and "lm\_head". No bias configuration was used, and the task type was Causal Language Modeling (CAUSAL\_LM). The learning rate for this model was set to 2e-4, and the batch size was 8. Optimization was carried out using the PagedAdamW 32-bit optimizer.

\paragraph{GPT-35 and GPT-4}
They have been fine-tuned using LoRa on default settings used for these models in Azure API documentation\footnote{\url{https://learn.microsoft.com/en-us/azure/ai-services/openai/}}.
\section{Technical Details of validators} \label{subsec:details}
In this section, we provide the details about the prompts used with each validator in the above study. We use greedy decoding for all prompts to ensure more precise computation.

\begin{figure}[h]
    \centering
    \includegraphics[width=\columnwidth]{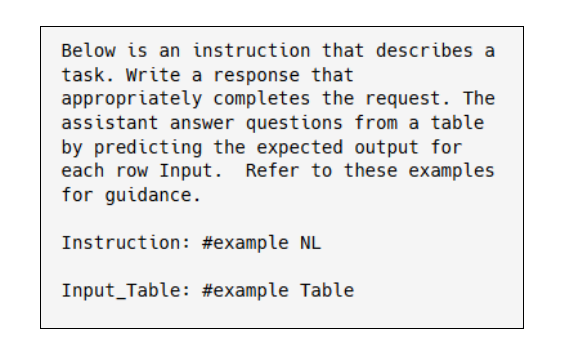}
    \caption{Prompt used for Output Prediction validation}
    \label{prompt_DC}
\end{figure}

\paragraph{Output Prediction}
We use an LLM (GPT-4) as a validator V. We prompt the LLM with input table and NL to compute the output for the target column directly. Then we validate the target output by comparing them with the actual outputs, with validation deemed successful only if the expected and actual outputs match for all rows in the table. The matching criteria differs based on the datatype: for a numeric value we allow an absolute difference of up to 0.05 and a string is considered a match when the longest matching contiguous sub-sequence coefficient (defined as length of longest matching sub-sequence divided by length of the longer string) is greater than 0.8.
The prompt used for this technique is provided in Figure \ref{prompt_DC}.



\begin{figure}[h]
    \centering
    \includegraphics[width=\columnwidth]{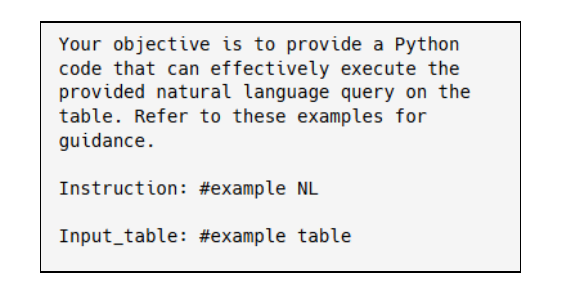}
    \caption{Prompt used for Alternate code generation validation}
    \label{prompt_CE}
\end{figure}

\paragraph{Alternate code generation}
We use an LLM (GPT-4) as V and task it with Python generation using NL and table as the input. The matching criterion is same as that of Direct computation.
The prompt is provided in Figure \ref{prompt_CE}.
 
 
 
\begin{figure}[h]
    \centering
    \includegraphics[width=\columnwidth]{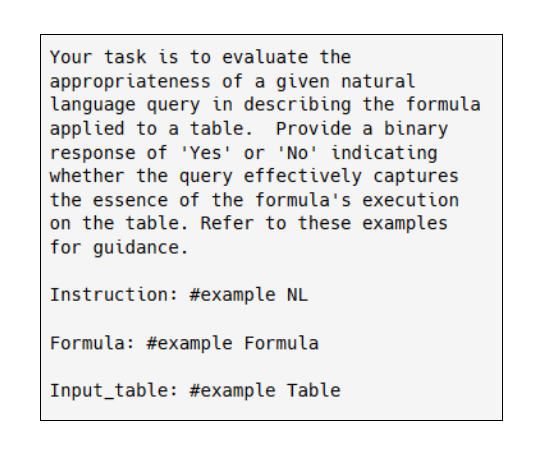}
    \caption{Prompt used for Classification validation}
    \label{prompt_C}
\end{figure}

\paragraph{Classification}
We prompt an LLM (GPT-4) as validator V to generate a binary outcome, judging whether the given natural language query accurately describes the formula when applied to the corresponding table.
The prompt is provided in Figure \ref{prompt_C}.
 
 
 
\end{document}